\documentclass[10pt,twocolumn,letterpaper]{article}

\usepackage{wacv}
\usepackage{times}
\usepackage{epsfig}
\usepackage{graphicx}
\usepackage{amsmath}
\usepackage{amssymb}
\usepackage{multirow}
\usepackage{hyperref}

\wacvfinalcopy 

\def\wacvPaperID{64} 

\ifwacvfinal\fi
\begin{document}

\title{Fixation prediction with a combined model of bottom-up saliency\\ and vanishing point}

\author{Mengyang Feng$^{+}$  \ \ \ \ \ Ali Borji$^{\dagger}$ \ \ \ \ \ Huchuan Lu$^{+}$ \\
$^{\dagger}$Computer Science Department, University of Wisconsin - Milwaukee, USA\\
$^{+}$Department of Electrical Engineering, Dalian University of Technology, China\\
{\url{archerfmy@163.com, borji@uwm.edu, luhuchuan@gmail.com}}}

\maketitle
\ifwacvfinal\thispagestyle{empty}\fi

\begin{abstract}

\vspace{-10pt}
By predicting where humans look in natural scenes, we can understand how they perceive complex natural scenes and prioritize information for further high-level visual processing. Several models have been proposed for this purpose, yet there is a gap between best existing saliency models and human performance. While many researchers have developed purely computational models for fixation prediction, less attempts have been made to discover cognitive factors that guide gaze. Here, we study the effect of a particular type of scene structural information, known as the vanishing point, and show that human gaze is attracted to the vanishing point regions. We record eye movements of 10 observers over 532 images, out of which 319 have vanishing points. We then construct a combined model of traditional saliency and a vanishing point channel and show that our model outperforms state of the art saliency models using three scores on our dataset. 

\vspace{-10pt}
\end{abstract}
\vspace{-5pt}
\section{Introduction}
\label{sec:intro}



Visual attention and eye movements are crucial in understanding complex scenes. 
Primates use focal visual attention and rapid eye movements
to analyze complex visual inputs in real-time, in a manner that highly
depends on current behavioral priorities and goals. Yarbus (1967)~\cite{yarbus1967eye} demonstrated a striking example of how a
verbally-communicated task specification may dramatically affect attentional
deployment and eye movements. He argued that variable spatiotemporal
characteristics of scanpath for different task specifications
exemplify the extent to which behavioral goals may affect eye movements and
scene analysis. Some eye tracking experiments in the context of spoken sentence comprehension
have shown that the interplay between task demands
and active vision is reciprocal. For example, Tanenhaus et al. (1995)~\cite{tanenhaus1995integration} tracked
fixations of subjects when they received ambiguous verbal instructions regarding
manipulating objects on a table. Tanenhaus et al. demonstrated that visual context
influenced spoken word recognition and syntactic processing when subjects had to resolve 
ambiguous verbal instructions by analyzing the visual scene and objects. These two studies 
indicate that visul attention and scene understanding are intimately interrelated.


Following two seminal works, Feature Integration Theory by Triesman and Gelade~\cite{TriesmanGelade} and the computational attention architecture by Koch and Ullman~\cite{koch1987shifts}, several attention models have been proposed to detect bottom-up salient regions that stand out from their surroundings in an image~\cite{borji2013state,borji2013quantitative,judd2012benchmark}. These models can be classified under three categories: 1) purely computational, 2) purely cognitive, and 3) a hybrid of both computational and cognitive. Models in the first category intend to detect salient regions often by using machine learning or statistical tools. For example, some researchers have formulated the problem as a classification problem by trying to estimate which points (and to what degree) will be looked at by humans (e.g.,~\cite{harel2007graph,hou2007saliency,zhang2013saliency,judd2009learning,bruce2005saliency}). Some studies in the second category have been investigating cognitive factors that influence eye movements in free viewing of natural scenes. These are often behavioral studies that accurately formulate/analyse hypotheses and rule out confounding factors.
For example, it has been shown that eye movements are driven to the center of objects~\cite{nuthmann2010object} and scenes~\cite{tatler2005visual} or gaze direction of actors in scenes direct viewers' gaze~\cite{borji2014complementary,parks2014augmented}. Models in the third category are either inspired by the mechanisms of human attention and mimic it (e.g.,~\cite{1998Itti,GarciaDiazJOV}) and/or have used a set of cognitive factors to build a model to predict fixations (e.g.,~\cite{cerf2007predicting,parkhurst2002modeling}). Note that some studies and models fall under more than one category and the categories are not exclusive.

Several cognitive cues that attract attention and guide eye movements have already been discovered (e.g., color, texture, motion, text~\cite{cerf2007predicting}, face~\cite{cerf2007predicting}, object center-bias~\cite{nuthmann2010object}, scene center-bias~\cite{tatler2005visual}, cultural cues~\cite{chua2005cultural,rayner2009eye}, and gaze direction~\cite{castelhano2007see,borji2014complementary}). Scene structural information such as scene gist (global context), scene layout, horizontal line, depth, and openness influence eye movements as well as human scene categorization~\cite{torralba2006contextual}. Here, we systematically investigate the role of a particular type
of scene structural information, known as the vanishing point (VP) and perspective, on eye movements in free-viewing of pictures of natural scenes and propose a combined model of bottom-up saliency and VP. Therefore, our model is classified under the third category mentioned above.

%



In graphical perspective, a vanishing point is a 2D point
(in image plane) which is the intersection of parallel lines in
the 3D world (but not parallel to the image plane). VP can
be seen in fields, rail roads, streets, tunnels, forest, buildings,
objects such as ladder, etc. It has been used in camera
calibration, 3D reconstruction as well as in painting. 






\begin{figure}[t]
\begin{center}
\includegraphics[width=\linewidth]{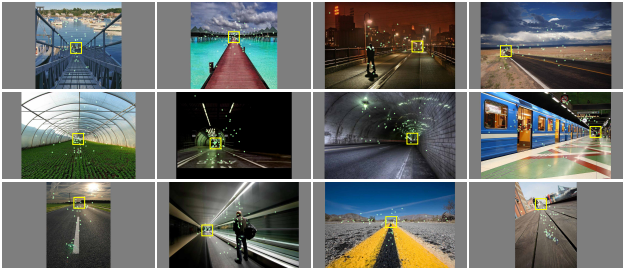}
\end{center}
\vspace{-10pt}
   \caption{Example stimuli with vanishing points (yellow boxes) and fixations (dots). For some images with highly salient items, the vanishing point attracts less attention (e.g., bottom-right image). For some images, salient content happens at the vanishing point while for some others it does not. 
}
\label{fig:stim}
\vspace{-10pt}
\end{figure}

\begin{figure*}[t]
\vspace{-5pt}
\begin{center}
\includegraphics[width=0.9\linewidth]{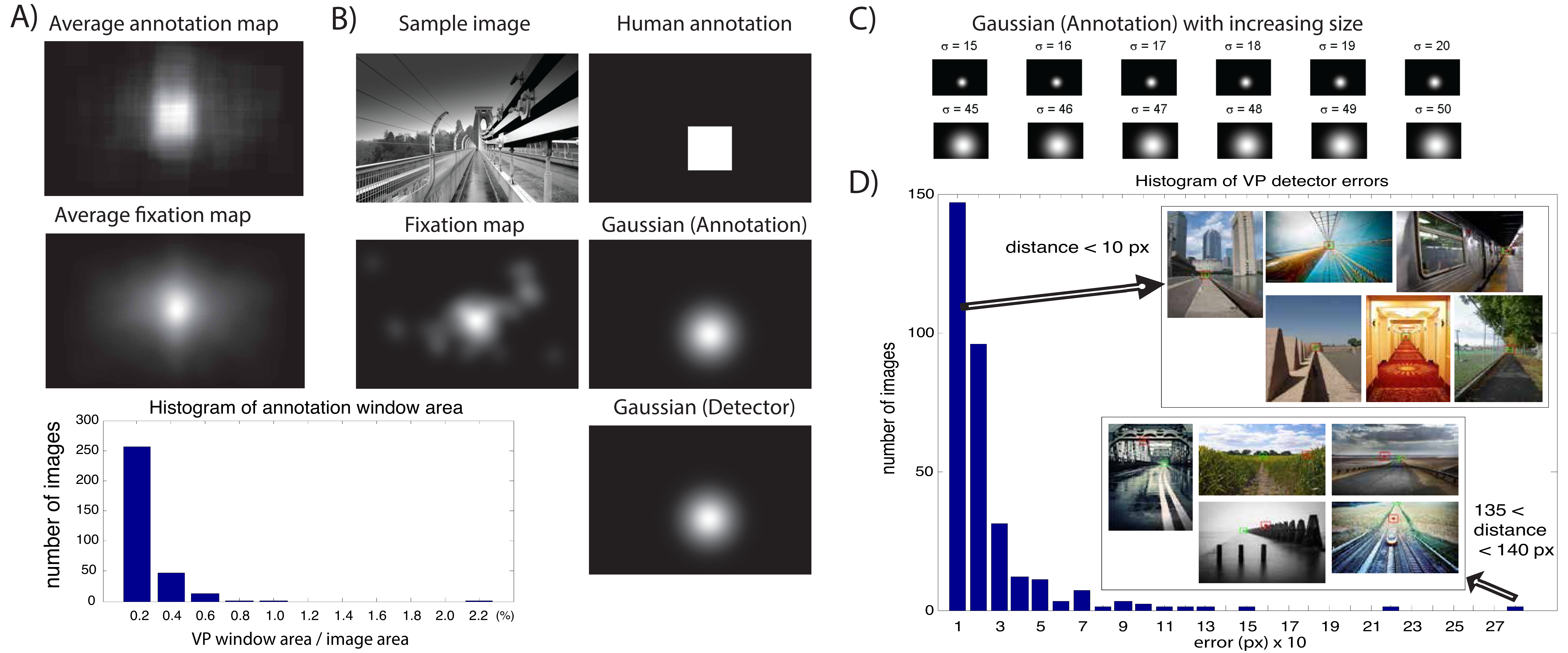}
\end{center}
\vspace{-10pt}
   \caption{\small{A) Average VP map and average fixation map over images with vanishing points. B) A sample image, its vanishing point map, eye movements, Gassian blob placed at the vanishing point, and Gaussian placed at the vanishing point using our automatic VP detector. C) Illustration of the Gaussian at the vanishing point with increasing $\sigma_{vp}$. D) Histogram of vanishing point detector errors. Bins here are [0,5), [5, 10), $\cdots$, and [135,140). Some examples when detector works well or fails are also shown (Green= Ground truth, Red= Detection).
}}
\label{fig:illust}
\vspace{-10pt}
\end{figure*}





%

\section{Data Collection}
\vspace{-2pt}
\subsection{Stimuli}
\vspace{-3pt}


Our stimuli contains 319 color images with vanishing points with resolution of 1920 $\times$ 1080 pixels (with added gray margins while preserving the aspect ratio) from different categories. Firstly, we collected 700 images from Google search, MIT300~\cite{judd2012benchmark} and DUT-OMRON~\cite{OMRON_DB} datasets. We ruled out images with more than one vanishing point and images with complex texture informations which may cost the disadvantage of (automatically) detecting the vanishing point. Eventually, we were left with 319 images and manually annotated vanishing points by drawing rectangles around them. Two members of our laboratory completed the annotation task together. Average height of VP rectangles is 10px and average width is 14px (only center of VP is used here). Since showing only images with a vanishing point may generate a viewing bias in observers and draw them automatically to vanishing points, we then gathered additional 213 images without vanishing points, and shuffled them among images with VPs. Therefore, viewers would not know in advance whether a presented image will have a VP.
We had 532 images in total to record human fixations, out of which 319 had VP and 213 did not. In our modeling and experiments here, we only analyze 319 images with VPs. Figure~\ref{fig:stim} shows examples of our stimuli, labeled vanishing points, as well as fixation locations.

Figure~\ref{fig:illust}.A shows average VP annotation map as well as average fixation map over 319 images with VPs. Both of these maps have maxima at the image center making center-bias a potential confounding factor which we will address extensively in our analyses. This figure also shows the histogram of VP window size. 82.25\% of VP rectangles have size smaller than 0.2\% the image area.


\subsection{Eye tracking}


\textbf{Observers:}  We had 10 observers (6 male, 4 female) in total. Mean observer age was 22 (min=21, max=24, median 22, std 0.84). Observers were undergraduates at our university from different majors and cities.
Observers had normal or corrected-to-normal vision and received course credit for participation. They were naive to the purpose of the experiment and had not previously seen the stimuli.


\textbf{Procedure:} 
Following the fixation cross, a target image was shown for 4 seconds followed by 3 seconds gray screen.
   Observers sat 60 cm away from a 19 inch LCD monitor screen such that scenes subtended approximately 37.6 degree $\times$ 24 degrees of visual angle. A chin rest was used to stabilize head movements. Stimuli were presented at 60Hz at a resolution of 1920 $\times$ 1080 pixels (with added gray margins while preserving the aspect ratio). Eye movements were recorded via a Tobii X1 Light Eye Tracker at a sample rate of 1000Hz. The eye tracker was calibrated using 5 points calibration at the beginning of each recording session.
Observers viewed images in random order.

\section{Our Model}

In this section, we present details of our learning model with human annotations first. We then mention how we will automatically detect VPs to replace human annotations. We also compare performance of our model with human annotations and with automatic detections.
	
\subsection{Learning a combined saliency map}


Each image pixel is represented by X = [s v] where s is the output of a bottom-up saliency model (e.g., AIM~\cite{bruce2005saliency}, BMS~\cite{zhang2013saliency}, and Itti~\cite{1998Itti}). v is the value from the vanishing point map (VP) modeled as a variable size Gaussian placed at the vanishing point as shown in Figures~\ref{fig:illust}.B \&~\ref{fig:illust}.C\footnote{We experimentally found that the Gaussian form of VP works better than a rectangle or a circle.}: 
\begin{equation}
VP(x,y) = \frac{1}{2 \pi \sigma _{vp} ^2} e^{-\frac{(x-i)^2 + (y-j)^2}{4 \sigma _{vp} ^2}}
\end{equation}

\noindent where $(i,j)$ is the coordinate of the annotated vanishing point and $\sigma_{vp}$ is the (variable) standard deviation of the Gaussian blob. In section~\ref{auto}, the coordinate of the vanishing point will be replaced by our automatic VP detector.

 We aim to learn f(X) = $W^{T}$X + $b$ which is a binary function determining whether location with feature vector X should be attended or not. To do so, we use a SVM with a linear kernel. For a test pixel, we assign the m = $W^{T}$X + $b$ as the label of it. Final saliency values are then normalized for each map (i.e., (m - min) / (max - min)).
We avoid using complicated non-linear learning functions (e.g., boosting) here deliberately, since we are interested to find out the real added value of the vanishing point.


We choose 50 random images for training the SVM and the rest 269 images for testing.
We randomly select 50 pixels respectively from fixated locations and non-fixated locations, yielding 100 samples (50 positive samples and 50 negative samples) for each training image, i.e., 5000 samples in total. 
Note that we cut off the added gray margins and resized the maximum length of the  image side to 400 pixels while preserving the aspect ratio (to reduce the calculation).


We learn the combined models (e.g., AIM + VP, BMS + VP, and Itti + VP) and compare them with the original bottom-up saliency models, respectively. 






\subsection{Automatic detection of vanishing points}
\label{auto}
Several methods for detecting vanishing points in an image exist (See~\cite{kong2009vanishing}). Some methods utilize line segments detected in an image. Some other approaches consider intensity gradients of the image pixel. There can be several vanishing points present in an image. Here, our aim is to detect the vanishing point that corresponds to the principal directions (lines) in a scene.

Our method also utilizes line segments to get the vanishing points. For an input image, we use the PB boundary detection algorithm~\cite{martin2004learning} to obtain the boundary map $B$. $B(i,j)$ gives the probability of a boundary at each pixel $(i,j)$. 
We then applied Hough Transform~\cite{hough1959machine} to detect line segments. Since the input of the Hough Transform should be a binary map, we turn $B$ into a binary map using an adaptive threshold,

\begin{equation}
{B_2}\left( {i,j} \right) = \left\{ {\begin{array}{*{20}{c}}
1&{B(i,j) \ge t}\\
0&{B(i,j) < t}
\end{array}} \right.
\end{equation}

\noindent where $t = 10 \times \frac{\sum B(i,j)}{heigh \times weight}$. In this work, height and width are the size of the $B$ map. $t$ is chosen by experience.
 Then the line segments map $L$ is computed as $L=Hough(B_2, \theta, lt)$ where $\theta$ is the angle of lines which could be detected, and $lt$ represents the threshold when choosing a line. In this work, we set $\theta=180^\circ$ that means lines from every direction can be detected. And $lt=60$, which means that lines which have more than 60 pixels on them can be detected. Note that, the parameters $t$ and $lt$ ensure that only the large line segments could be detected.

\begin{equation}
L(i,j) = \left\{ {\begin{array}{*{20}{c}}
1, &{(i,j) \in {l_d}}\\
0, &{otherwise}
\end{array}} \right.
\end{equation}

\noindent where $l_d$ presents the detected lines. From the line segments map $L(i,j)$, we can get the intersections of those lines. Our aim is to detect the vanishing point that corresponds to the principal directions in a scene. So, we can assume that the location $(X,Y)$ where most intersections happen around it could be the vanishing point. More specifically, if two intersections' Euclidean distance is smaller than 10 pixels, we consider that they are the neighbor points, and calculate the number of neighboring points around each intersection. Then, we can find the intersection $(i_v,j_v)$ which has most neighboring points around and calculate the VP location $(X,Y)$ using this formula,
\begin{equation}
X = \frac{\sum_1^{M} x_i}{M}; \ \ Y = \frac{\sum_1^{M} y_i}{M}; \ \ i = 1, 2, \cdots, M 
\end{equation}

\noindent where $M$ is the number of the neighbor points around $(i_v, j_v)$, and $(x_i, y_i)$ presents the coordinate of the ith neighbor. The method performed well over our dataset.


\section{Experiments and Results}
\label{Res}




 

Firstly, we aim to optimize our combined model by finding the best $\sigma_{vp}$. Table~\ref{tab:scores} summarizes the results by reporting the point where performance is maximum. $\sigma_{vp}$ is changing from $\sigma_{vp}=15$ pixel to $\sigma_{vp}=50$ pixel.
We observe more than 10\% improvement of Model + VP versus Model using the AUC scores with any of the three models. Improvement using NSS is more than 50\% while improvement using CC is more than 60\%. Considering both NSS and CC scores, we determined the $\sigma_{vp}^{best}$ by adding the normalized NSS and CC scores and selecting the $\sigma_{vp}$ value corresponding to the peak. Then Model + VP$_b$ represents the optimized combined model.

\begin{figure}[t]
\begin{center}
\includegraphics[width=\linewidth]{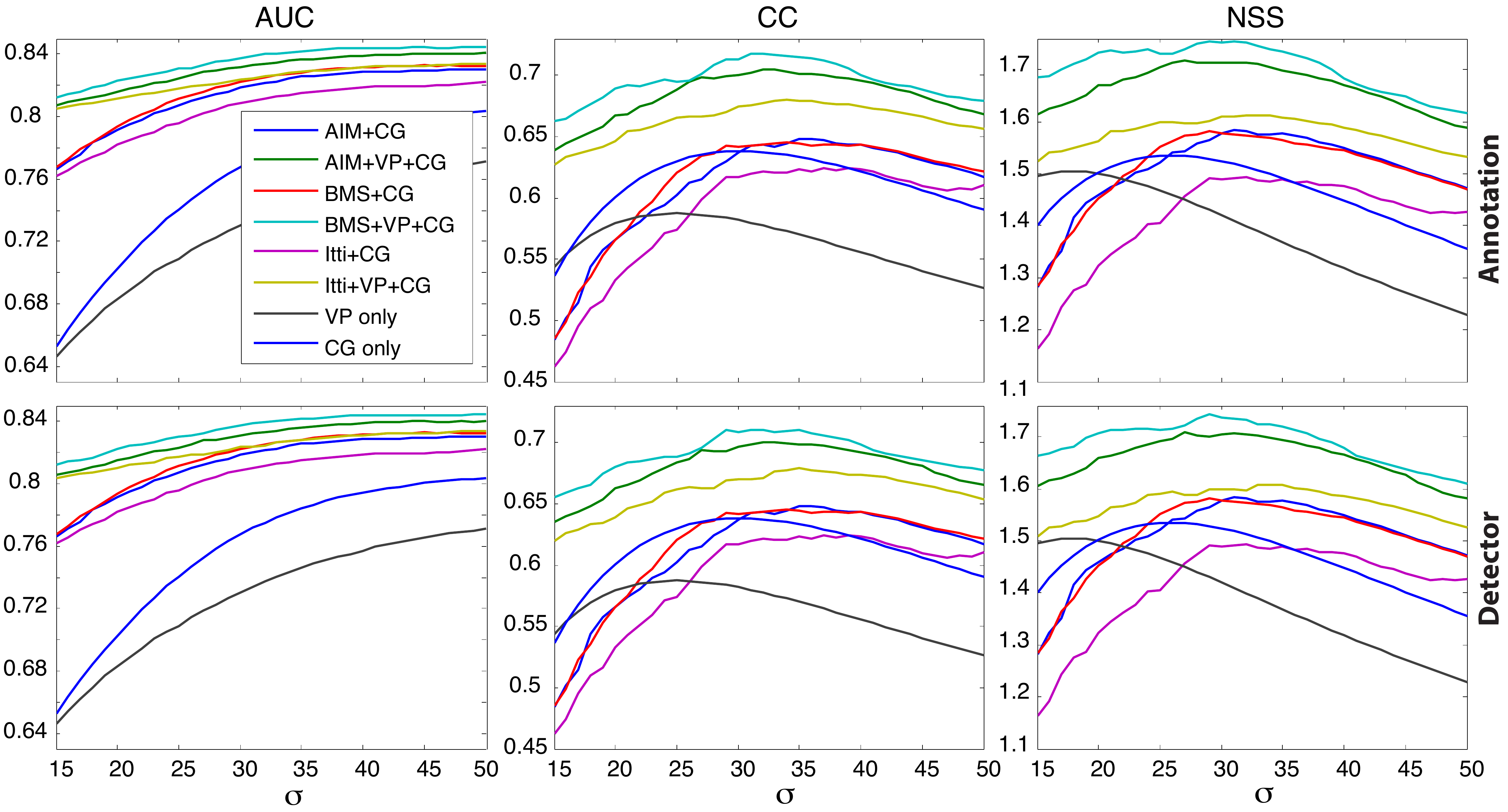}
\end{center}
\vspace{-10pt}
   \caption{Performance of models using AUC, CC, and NSS scores as a function of $\sigma_{cg}$ in Model + VP$_b$ + CG. The top row shows model performance using the human annotation and the bottom row shows the performance using the automatic VP detector. As you can see performance using detector and human annotations are very close to each other. Since the detector error is small relative to Gaussian size, it gets canceled.
}
\label{fig:scoresCG}
\vspace{-13pt}
\end{figure}

Since center-bias~\cite{tatler2005visual} is an important confounding factor, here we compare the Model + VP$_b$ + CG(Central Gaussian map) to Model + CG to see whether the VP is the main cause or not.
Figure~\ref{fig:scoresCG} shows scores of models as a function of $\sigma _{cg}$(i.e., the $\sigma$ of the Central Gaussian). As this figure shows increasing $\sigma _{cg}$ increases the AUC score until it saturates. Performance peaks using NSS~\cite{peters2005components} and CC and then declines. Model + CG works better than CG and VP only maps but performs below Model + VP$_b$ + CG. This trend happens using all three scores but is more prevalent using CC and NSS scores. Interestingly, performance using our VP detector is very close to the performance using human annotations (although slightly lower). The automatic VP detector has error in locating the correct location of the VP for some images, but this error is negligible because it does not affect the placement of the Gaussian blob (i.e., smoothing). In both rows, baseline models score below all shown models including VP only and CG only models.

\begin{table}[b]
\vspace{-15pt}
\begin{center}
\footnotesize{
  \renewcommand{\arraystretch}{1}
  \renewcommand{\tabcolsep}{.4mm}

\begin{tabular}{|l|c|c|c|c|c|c|c||c|}
\hline
Score &  &  AIM & AIM + VP & BMS & BMS + VP & Itti & Itti + VP & VP only \\
\hline\hline

\multirow{3}{*}{AUC}  & A &  0.719 & 0.807 (50) & 0.708 & 0.812 (50) & 0.727 & 0.801 (50) & 0.771 (50) \\
& D   & 0.719 & 0.804 (50) & 0.708 & 0.809 (50) & 0.727 & 0.798 (50) & 0.771 (50) \\
& I   & - & 12.2\% & - & 14.7\% & - & 10.2\% & -  \\
& W   & - & [5.8, 6.5] & - & [7.1, 6.1] & - & [5.1, 6.0] &  \\
\hline\hline

\multirow{3}{*}{NSS} &A &  0.957 & 1.495 (31) & 0.916 & 1.544 (26) & 0.903 & 1.412 (37) & 1.505 (18) \\
& D &  0.957 & 1.483 (27) & 0.916 & 1.512 (26) & 0.903 &1.397 (37) & 1.505 (18) \\
& I   & - & 56.2\% & - & 68.6\% & - & 56.4\% &  \\
& W   & - & [5.8, 6.5] & - & [7.1, 6.1] & - & [5.1, 6.0] &  \\

\hline\hline

\multirow{3}{*}{CC} & A &  0.356&0.601 (38)&0.340&0.612 (33)&0.362&0.585 (37)&0.587 (25) \\
& D & 0.356& 0.591 (36) &0.340 &0.600 (33)&0.362 & 0.578 (37) & 0.587 (25) \\
& I  & - & 68.6\% & - & 80\% & - & 61.6\% & - \\
& W   &  - & [8.0, 6.9] & - & [9.5, 6.7] & - & [6.9, 6.9] &  \\
\hline
\end{tabular}
}
\end{center}
\caption{Scores of our combined model (Model + VP) vs. the original model and VP only channel. Numbers in parentheses are the best $\sigma _{vp}$ where performance peaks using different scores. Improvement is measured with respect to the original model. W is the parameters of the line learned by the SVM classifier. $\sigma _{vp}^{best}$ using human annotations and auto detector for AIM, BMS, and Itti models in order are 31, 27, and 37 pixels. (A= Annotation, D= Detection, I= Improvement, W= Weights).}
\label{tab:scores}
\end{table}

\begin{table*}[t]
\vspace{-5pt}
\begin{center}
\footnotesize{
  \renewcommand{\arraystretch}{1}
  \renewcommand{\tabcolsep}{2mm}

\begin{tabular}{|l|c|c|c|c|c|c|c||c|}
\hline
Score &  &  AIM+C & AIM+C+V$_b$ & BMS+C  & BMS+C+V$_b$ & Itti+C & Itti+C+V$_b$ & C only \\
\hline\hline

\multirow{3}{*}{AUC}  & Annotation (A) &  0.803 (48) & 0.841 (50) & 0.833 (45) & 0.845 (45) & 0.822 (50)& 0.834 (50) & 0.804 (50) \\
& Detection (D)  &  0.803 (48) & 0.840 (45) & 0.833 (45) & 0.845 (49) & 0.822 (50)& 0.834 (49) & 0.804 (50) \\
& Improvement (I)   & - & 4.7\% & - & 1.4\% & - & 1.5\% & - \\
& Weights (W)   &  [4.5,4.9] & [4.1,2.0,4.6] & [5.9,5.4] & [5.5,2.3,4.8] & [4.1, 5.4] & [3.7,2.5,4.9] & - \\
\hline\hline

\multirow{3}{*}{NSS}  & A &  1.584(31) & 1.719(27) & 1.582(29) & 1.755(29) & 1.493(32) & 1.613(34) & 1.535(28)\\
& D  & 1.584(31) & 1.709(27) & 1.582(29) & 1.745(29) & 1.493(32) & 1.608(35) & 1.535(25) \\
& I   & - & 8.5\% & - & 10.9\% & - & 8.0\% & - \\
& W   & [7.7, 6.9] & [7.2, 4.3,5.3] & [9.2,7.2] & [8.2,4.3,5.2] & [6.2,6] & [4.9,4.2,4.8] & - \\
\hline\hline

\multirow{3}{*}{CC}  & A &  0.648(35) & 0.705(33) & 0.645(34) & 0.718(32) & 0.624(37) & 0.680(34) & 0.638(30) \\
& D   & 0.648(35) & 0.700(33) & 0.645(34) & 0.711(32) & 0.624(37) & 0.679(35) & 0.638(30) \\
& I   & - & 8.8\% & - & 11.3\% & - & 9.0\% & - \\
& W   & [6.6,6.5] & [6.1,3.6,5.4] & [8.2,6.3] & [7.3,3.9,5.5] & [5.6,5.7] & [4.9,4.2,4.8] & - \\
\hline
\end{tabular}
}
\end{center}
\caption{Performance of models using AUC, CC, and NSS scores as a function of $\sigma _{cg}$ in Model + VP$_b$ + CG.}
\label{tab:scoresCG}
\vspace{-10pt}
\end{table*}

\begin{figure*}[htbp]
\begin{center}
\includegraphics[width=\linewidth]{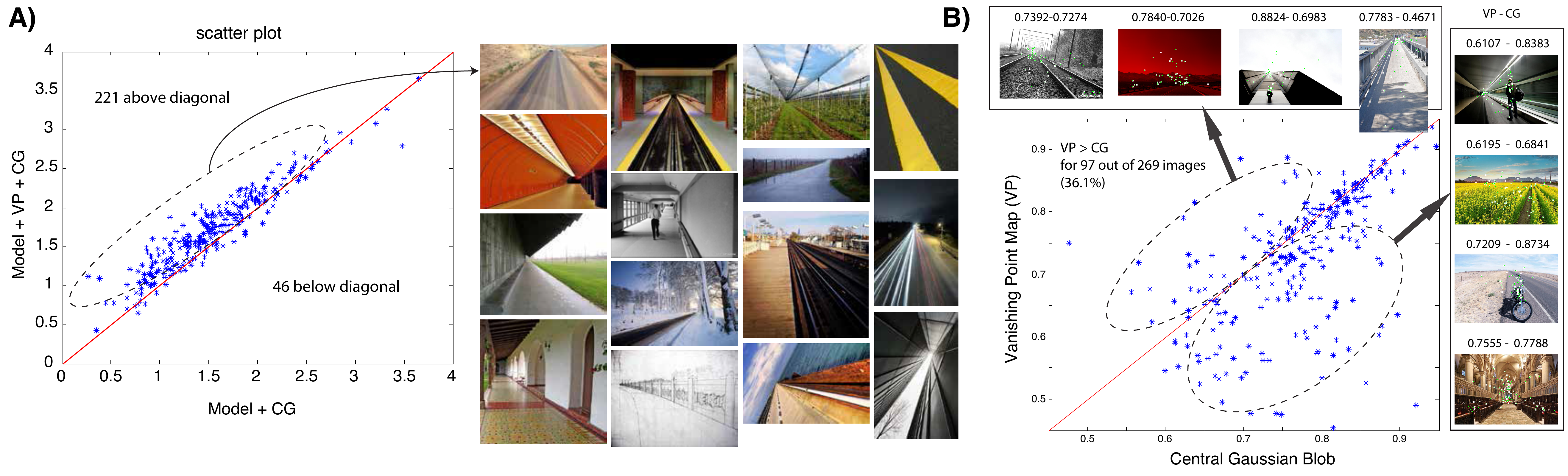}
\end{center}
\vspace{-15pt}
   \caption{A) Scatter plot of Model + CG versus Model + VP + CG using NSS score. Each dot represents one image. CG stands for Central Gaussian. This plot shows the added value of VP over the original model taking into account the center bias confound. So we can make sure that added value is not because of the center-bias. For 221 of images, we observe an improvement. 
Vanishing points usually happens off center on these images.
We did the same analysis by plotting the Model + VP versus Model and observed that for 243 of images performance is improved. B) AUC score of the VP map versus Central Gaussian map for $\sigma_{vp}=\sigma_{cg}=31$ pixels. Some examples above/below diagonal are shown.
}
\label{fig:histplotNew}
\vspace{-10pt}
\end{figure*}

To investigate the effect of center-bias on our results we have conducted two analyses shown in Figure~\ref{fig:histplotNew}. Figure~\ref{fig:histplotNew}.A shows performance of the center-bias modulated VP-added model versus center-bias modulated model for each image. For 221 of the images, we observe an improvement of the former over the latter. In fact, when observing these 221 images, we found that for the majority of the images, VP happened off the center. This means that adding VP increases the results additional to what adding center-bias offers. In Figure\ref{fig:histplotNew}.B, we plot the AUC score of the vanishing point (VP) map versus the Central Gaussian map (CG). As this figure shows, VP map wins over the Central Gaussian map for some images (36.1\% of test images). For many other images, VP happens at the image center. Thus, we conclude that vanishing point and central bias are two different phenomena with distinct effects, although over our stimuli they coincide in many images by construction.  

 Tables~\ref{tab:scoresCG} compares Model + CG versus Model + CG + VP$_b$ which addresses center-bias. 
Improvement of Model + VP$_b$ + CG over Model + CG is smaller using 3 scores (about 2.5\% using AUC average over 3 models, about 9.1\% using NSS, and about 9.7\% using CC). Investigating the parameters of the discriminant line learned by SVM (i.e., weight of the BU and VP) shows that both baseline model and VP map are involved in the final combination.

\begin{table}[htbp!]
\begin{center}
\footnotesize{
  \renewcommand{\arraystretch}{1}
  \renewcommand{\tabcolsep}{.2mm}

\begin{tabular}{|l|l|c|c|c|c|c|}
\hline
Score & Model & M + VP$_b$ + CG$_b$   & M + VP$_b$  &  M + VP$_b$ & VP$_b$ vs. \\
          &            &  vs. M + CG$_b$             &  vs. VP$_b$   &  vs. M            & Chance \\
\hline\hline
\multirow{6}{*}{AUC} & AIM & 0.833 vs. 0.819  &0.793 vs. 0.739 &0.793 vs. 0.720  & 0.739 vs.  0.5 \\
& & p=1.574e-24 &p=6.866e-30 &p=4.193e-35 & p=4.905e-40 \\
\cline{2-6}

 & BMS & 0.837 vs. 0.823  &0.798 vs. 0.719 &0.798 vs. 0.711  & 0.719 vs.  0.5\\
& & p=3.993e-21 &p=4.466e-32 &p=6.759e-34 & p=1.161e-39 \\
\cline{2-6}

 & Itti & 0.826 vs. 0.811  &0.792 vs. 0.756 &0.792 vs. 0.730  & 0.756 vs.  0.5\\
& & p=5.332e-21 &p=2.474e-22 &p=1.075e-30 & p=5.840e-40 \\

\hline\hline

\multirow{6}{*}{NSS} & AIM &1.695 vs. 1.545  &1.467 vs. 1.450 &1.467 vs. 0.953  & 1.450 vs. 0 \\
& & p=4.694e-21 &p=1.932e-03 &p=1.704e-28 & p=6.875e-39 \\
\cline{2-6}

 & BMS & 1.751 vs. 1.587  &1.543 vs. 1.508 &1.543 vs. 0.916  & 1.508 vs.  0 \\
& & p=7.277e-22 &p=5.273e-06 &p=6.738e-31 & p=4.217e-38 \\
\cline{2-6}

 & Itti & 1.592 vs. 1.445  &1.361 vs. 1.381 &1.361 vs. 0.918  & 1.381 vs.  0\\
& & p=1.384e-17 &p=7.499e-03 &p=2.091e-25 & p=1.370e-39 \\

\hline\hline
\multirow{6}{*}{CC} & AIM & 0.697 vs. 0.628  &0.584 vs. 0.598 &0.584 vs. 0.358  & 0.598 \\
& & p=6.328e-20 &p=4.755e-06 &p=5.343e-28 & -  \\
\cline{2-6}

 & BMS & 0.720 vs. 0.652  &0.609 vs. 0.608 &0.609 vs. 0.341  & 0.608  \\
& & p=4.673e-21 &p=8.347e-01 &p=2.709e-30 & - \\
\cline{2-6}

 & Itti & 0.672 vs. 0.603  &0.563 vs. 0.580 &0.563 vs. 0.369  & 0.580 \\
& & p=2.279e-17 &p=3.256e-05 &p=3.823e-25 & - \\

\hline
\end{tabular}
}
\end{center}
\caption{Statistical analysis of results and model comparison.}
\label{tab:test}
\vspace{-15pt}
\end{table}



To check the statistical significance of our results, we perform cross validation by randomly splitting data into two parts (50 train and 269 test). We train our SVM model on the train set and apply it to the test set. We repeat this procedure 20 times and compare the means. Results of statistical tests using t-test are shown in Table~\ref{tab:test}. 
We perform four comparisons: 1) M + VP$_b$ + CG$_b$(Central Gaussian using optimized $\sigma_{cg}$, similar to the $\sigma_{vp}^{best}$) vs. M + CG$_b$, 2) M + VP$_b$ vs. VP$_b$, 3) M + VP$_b$ vs. M, and 4) VP$_b$ vs. Chance. We find that VP model performs significantly above the chance level in predicting fixations. Adding VP to models significantly outperforms both VP and baseline models (taken individually). Tackling center-bias, by adding CG$_b$ to Model and to Model + VP$_b$, shows that VP is a significant cue in guiding gaze, independent of center-bias.
We obtain the same pattern of results using NSS and CC scores.

\begin{figure*}[t]
\vspace{-10pt}
\begin{center}
\includegraphics[width=\linewidth]{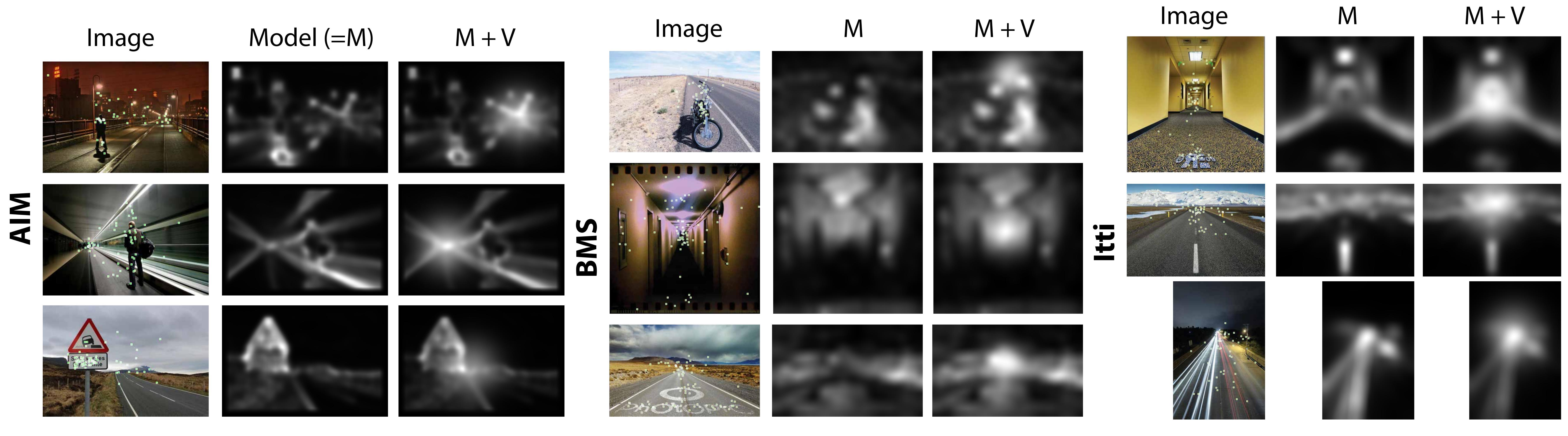}
\end{center}
\vspace{-10pt}
   \caption{Cases where our combined model performed poorly. In almost all of these cases, an object off the vanishing point overrides the VP and attract fixations.}
\label{fig:Bad}
\vspace{-10pt}
\end{figure*}

\begin{figure*}[t]
\vspace{-10pt}
\begin{center}
\includegraphics[width=.8\linewidth]{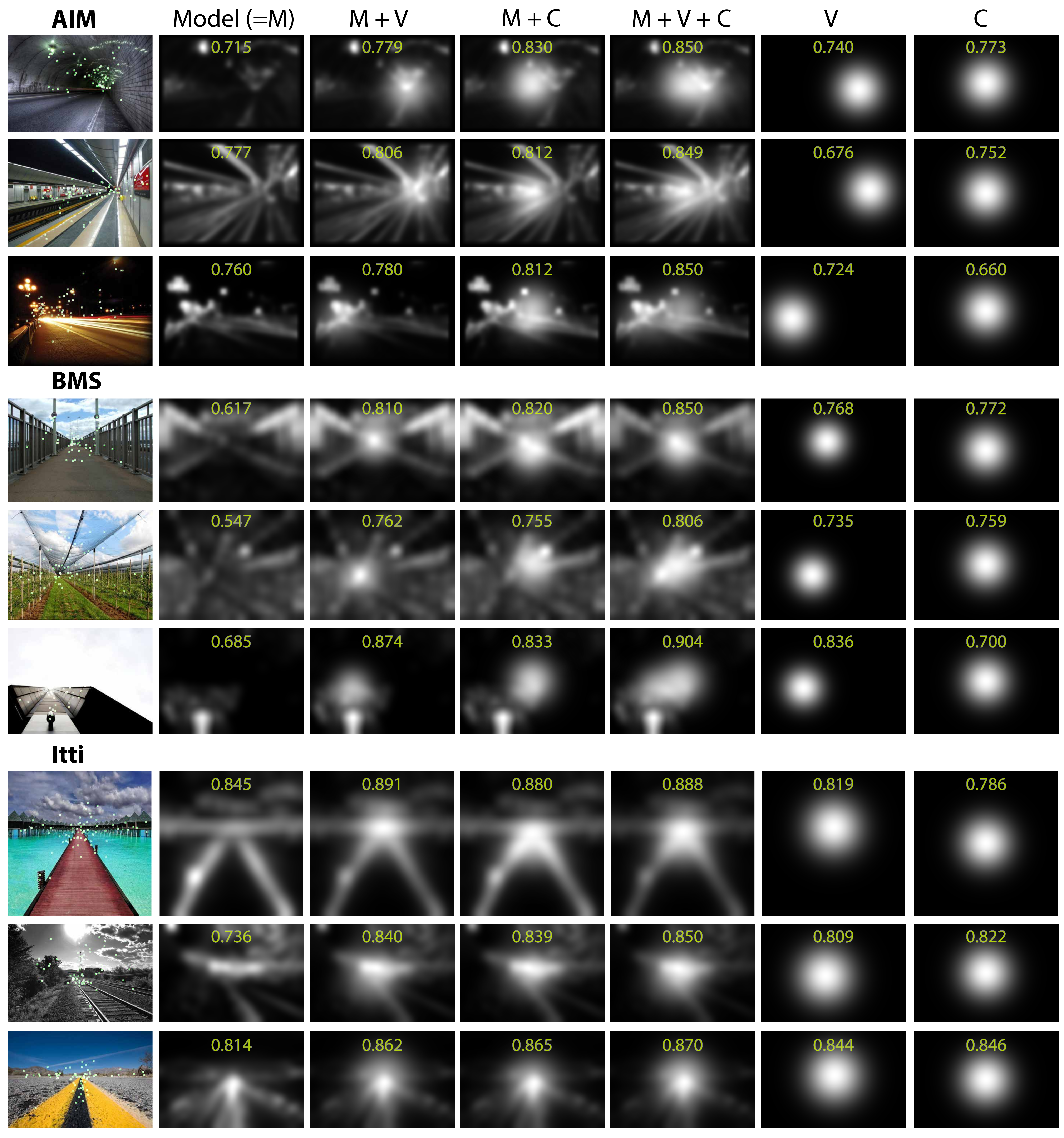}
\end{center}
\vspace{-10pt}
   \caption{Cases where our combined model performed well (using AUC score). }
\label{fig:Good}
\vspace{-10pt}
\end{figure*}

Figure~\ref{fig:Bad} shows examples where our combined model fails (i.e., M + VP$_b$ scores lower than M). In almost all these cases, an object off the vanishing point overrides the VP effect. Figure~\ref{fig:Good} shows examples where our model performs well along with scores of models.




\section{Discussion and Conclusion}
\vspace{-5pt}

We showed that vanishing point is a strong predictor of fixations in free viewing task and proposed a combined model of bottom-up saliency (using three state of the art models) and VP. Our model outperforms baseline models significantly with and without center-bias using three scores. We also showed that VP map performs significantly above chance. Since VP happens commonly in real life when taking pictures, we believe that adding it to models can in general enhance fixation prediction power. 

We intend to study the followings in future: 1) Whether (and to what extent) people prioritize vanishing points in presence of other salient cues in a scene? 2) Here, we added VP channel to images with a vanishing point. While this was not a problem with annotations, ultimately, we would like to add VP to only those images which have VP. For this we should automatically decide whether an image has VP or not (i.e., How much false positives of our detector will hurt?). In this regard, we will also consider images with multiple VPs, 3) We aim to 
relate our findings to other cues that might influence fixations in a similar fashion (but independently), for cues such as "focus of expansion" or "tangent line"~\cite{land1994we,land2001steering} and 4) we will consider other methods for detecting vanishing points in images (e.g., using convolutional neural networks~\cite{lecun1998gradient}).

We will share our dataset for further investigation of the role of the vanishing point cue in guiding gaze in free viewing. Hopefully our work will encourage more research toward discovering behavioral cues that guide attention and gaze in spatial and spatio-temporal domains.

{\small
\bibliographystyle{ieee}
\bibliography{egbib}
}
\end{document}